\pdfoutput=1

\documentclass[11pt]{article}

\usepackage{acl}

\usepackage{times}
\usepackage{latexsym}
\usepackage{graphicx}

\usepackage[T1]{fontenc}

\usepackage[utf8]{inputenc}

\usepackage{microtype}

\title{Multi-party Goal Tracking with LLMs: Comparing Pre-training, Fine-tuning, and Prompt Engineering}

\author{Angus Addlesee, \\ {\bf Weronika Siei\'nska}, {\bf Nancie Gunson}, \\ {\bf Daniel Hernandez Garcia}, {\bf Christian Dondrup} \\
  Heriot-Watt University, Edinburgh \\
  \texttt{\{a.addlesee, w.sieinska, n.gunson,} \\
  \texttt{d.hernandez\_garcia, c.dondrup\}@hw.ac.uk} \And Oliver Lemon \\
  Heriot-Watt University, Edinburgh \\
  Edinburgh Centre for Robotics \\
   Alana AI \\
  \texttt{o.lemon@hw.ac.uk} \\}

\begin{document}
\maketitle
\begin{abstract}
This paper evaluates the extent to which current Large Language Models (LLMs) can capture task-oriented multi-party conversations (MPCs). We have recorded and transcribed 29 MPCs between patients, their companions, and a social robot in a hospital. We then annotated this corpus for multi-party goal-tracking and intent-slot recognition. People share goals, answer each other's goals, and provide other people's goals in MPCs -- none of which occur in dyadic interactions. To understand user goals in MPCs, we compared three methods in zero-shot and few-shot settings: we fine-tuned T5, created pre-training tasks to train DialogLM using LED, and employed prompt engineering techniques with GPT-3.5-turbo, to determine which approach can complete this novel task with limited data. GPT-3.5-turbo significantly outperformed the others in a few-shot setting. The `reasoning' style prompt, when given 7\% of the corpus as example annotated conversations, was the best performing method. It correctly annotated 62.32\% of the goal tracking MPCs, and 69.57\% of the intent-slot recognition MPCs. A `story' style prompt increased model hallucination, which could be detrimental if deployed in safety-critical settings. We conclude that multi-party conversations still challenge state-of-the-art LLMs.
\end{abstract}

\section{Introduction}
\label{sec:intro}

Spoken Dialogue Systems (SDSs) are increasingly being embedded in social robots that are expected to seamlessly interact with people in populated public spaces like museums, airports, shopping centres, or hospital waiting rooms \cite{foster2019mummer,tian2021redesigning,gunson2022visually}. Unlike virtual agents or voice assistants (e.g. Alexa, Siri, or Google Assistant), which typically have dyadic interactions with a single user, social robots are often approached by pairs and groups of individuals \cite{al2012furhat,moujahid2022multi}. Families may approach a social robot in a\nobreakspace museum, and patients are often accompanied by a family member when visiting a hospital. In these multi-party scenarios, tasks that are considered trivial for SDSs become substantially more complex \cite{traum2004issues,zhong2022dialoglm,addlesee2023data}. In multi-party conversations (MPCs), the social robot must determine which user said an utterance,  who that utterance was directed to, when to respond, and what it should say depending on whom the robot is addressing \cite{hu2019gsn,gu2021mpc,gu2022hetermpc}. These tasks are collectively referred to as ``who says what to whom'' in the multi-party literature \cite{gu2022says}, but these tasks alone provide no incentive for a system to actually help a user reach their goals. State of the art ``who says what to whom'' systems can, therefore, only mimic what a good MPC \emph{looks like} \cite{addlesee2023data}, but for practical systems we also need to know what each user's goals are. We therefore propose two further tasks that become substantially more complex when considered in a multi-party setting: goal tracking and intent-slot recognition \cite{addlesee2023data}.

\begin{table}
\centering
\begin{tabular}{lll}
\hline
1 & U1: & What time was our appointment?       \\
2 & U2: & We have an appointment at 10.30pm.       \\
3 & U1: & Ok.                              \\ \hline
\end{tabular}
\caption{An example extract from our new corpus. This example illustrates that people complete other user's goals in an MPC. The system must understand that U1's question was answered by U2, and it does not need to answer this question as if it was a dyadic interaction. Further annotated examples can be found in Table \ref{tab:ex3}.}
\label{tab:ex1}
\end{table}

Dialogue State Tracking (DST) is a well-established task \cite{lee2021dialogue,feng2022dynamic} that is considered crucial to the success of a dialogue system \cite{williams2016dialog}. DST corpora are abundant \cite{henderson2014second,henderson2014third}, but they only contain dyadic conversations. No corpus exists containing MPCs with goal tracking or intent-slot annotations, yet there are important differences. Consider the example in Table \ref{tab:ex1} (from our new corpus, detailed in Section \ref{sec:dataset}). In turn 1, we can identify that User 1 (U1) wants to know their appointment time. Before the social robot had time to answer, User 2 (U2) answered in turn 2. This obviously does not occur in a dyadic interaction, yet this understanding is essential for natural system behaviour. The SDS must determine that it should not repeat the answer to the question, so data must be collected to learn this. Other major differences exist too. For example, current DST corpora do not contain a concept of `shared goals' \cite{eshghi2016collective}. If two people approach a café counter, the barista must determine whether the two people are separate (two individuals wanting to get coffee), or together (two friends with the shared goal to get coffee) \cite{keizer2013training}. The interaction changes depending on this fact, it would be unusual to ask ``are you paying together'' to two individuals. Shared goals can commonly be identified through explicit dialogue. For example, the use of `we' in ``We are looking for the bathrooms''. Similar to answering each other's questions, people may also ask questions on behalf of others. In our corpus, a person said ``ARI, the person that I'm accompanying feels intimidated by you, and they'd like to know where they can eat''.

In this paper, we present several contributions. (1) We collected a corpus of multi-party interactions between a social robot and patients with their companions in a hospital memory clinic. (2) This corpus was annotated for the standard ``who says what to whom'' tasks, but also for multi-party goal tracking and intent-slot recognition. We followed current DST annotation instructions, tweaked to enable annotation of multi-party phenomena (detailed in Section \ref{sec:dataset}). (3) We then evaluated Large Language Models (LLMs) on these two new tasks using our collected corpus. Models were pre-trained, fine-tuned, or prompt engineered where applicable (detailed in Section \ref{sec:experiment}). It is not possible to collect enormous corpora from patients in a hospital, so models were evaluated in zero-shot and few-shot settings. We found that the GPT-3.5-turbo model significantly outperformed others on both tasks when given a `reasoning' style prompt. 

\section{Dataset and Tasks}
\label{sec:dataset}

For the initial data collection, we partnered with a hospital in Paris, France, and secured ethical approval as part of the EU SPRING project\footnote{\url{https://spring-h2020.eu/}}. We then recorded, transcribed, translated (from French to English), anonymised, and annotated 29 multi-party conversations (774 turns). These MPCs were between patients of the memory clinic, their companion (usually a family member), and a social humanoid robot created by PAL Robotics called ARI \cite{cooper2020ari}. We hired a professional translator to avoid machine translation errors, and to enable faster experimentation as we are not French speakers. Future work based upon the findings in this paper will be evaluated in both English and French.

\begin{figure}
    \centering
    \includegraphics[width=0.48\textwidth]{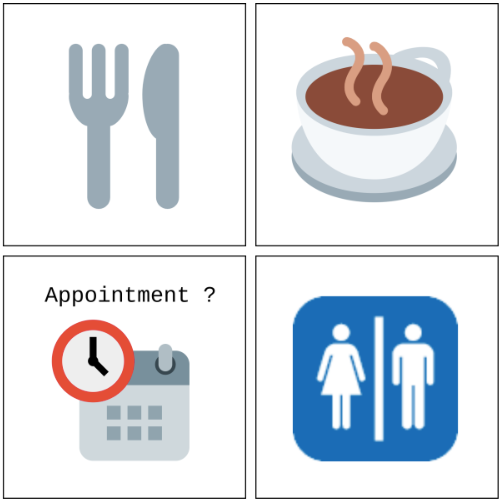}
    \caption{A sample of the pictograms  used to represent user goals, given to patients and companions. These elicited dialogues without restricting vocabulary.}
    \label{fig:pictogram}
\end{figure}

We used a wizard-of-oz setup as this task is new, and we required this data to design a multi-party SDS for use in the hospital. A robot operator was therefore controlling what ARI said by selecting one of 31 response options (task-specific answers and some common responses like ``yes'', ``no'', ``please'', ``thank you'', and ``I don't know''). Following our previously published data collection design \cite{addlesee2023data}, each participant was given one or two goals, and asked to converse with ARI to try to achieve their goal. Both participants were given the same goals in some cases to elicit dialogues containing `shared goal' behaviour. In order to encourage lexical diversity, we provided pictograms to give each participant their goals. For example, if we told the patient that they want a\nobreakspace latte, they would likely use the specific word ``latte'' \cite{novikova-etal-2016-crowd}, so we instead gave the participants   pictograms as seen in the top-right of Figure \ref{fig:pictogram}. This worked as people didn't just ask for coffee when given this image, some asked for hot chocolate or herbal tea instead.

In this paper, we evaluated each model on both multi-party goal tracking, and multi-party intent-slot recognition. These are two related, yet distinct tasks. If ARI asked the user ``Are you hungry?'', and the user responded ``yes'', then the intent of that turn is an affirmation, but the user's goal is also established as wanting to eat. As explained in Section \ref{sec:intro}, standard DST annotation schemes are designed for dyadic interactions, which do not enable annotation of multi-party behaviours. Each turn is annotated with its intent and slot values where applicable, but goal annotations require both the goal and the user whose goal is being established. When a goal is detected in a dyadic interaction, no user information is needed as there is only a single user. In multi-party interactions, multiple users can have multiple active goals. These goals may be different, they may be shared (see Table \ref{tab:ex2}), users may answer each other's goals (see Table \ref{tab:ex1}), and one user may provide another user's goal, for example by saying ``My wife would love a coffee''.

\begin{table*}
\resizebox{0.81\textwidth}{!}{\begin{minipage}{\textwidth}
\begin{tabular}{lllll}
\hline
  & \textbf{User} & \textbf{Utterance} & \textbf{Intent-Slot Annotation} & \textbf{Goal Tracking Annotation} \\ \hline
1 & U1:  & Hello, we'd like a coffee. Where can we go? & greet() ; request(beverage(coffee)) & G(U1+U2, drink(coffee))       \\
2 & ARI: & You have to enter the building behind you. & inform(directions(cafe)) & AG(U1+U2, drink(coffee))        \\
3 & U2:  & Ok, well thank you very much. & acknowledge(); thank() & CG(U1+U2, drink(coffee))                              \\ \hline
\end{tabular}
\caption{A corpus example displaying shared goals with both intent-slot and goal tracking annotations.}
\label{tab:ex2}
\end{minipage}}
\end{table*}

An annotated extract from an MPC in our collected corpus can be found in Table \ref{tab:ex2}. In turn 1, U1 states that ``we'd like a coffee'', indicating that U1 and their companion U2 would \emph{both} like a coffee. This turn is annotated with two intents: {\tt greet} (due to the ``hello''), and {\tt request}. This request intent has a slot value to indicate that the request is for a beverage -- coffee. The goal tracking annotation signifies that a goal has been established in this turn with `G'. The goal is shared by `U1+U2', and their goal is to drink a coffee. In turn 2, ARI responds informing both users where the café is, hence the {\tt inform} intent annotation. The goal tracking annotation is the same as turn 1, but starts with `AG' (for `answer-goal') instead of simply `G'. This indicates that this goal has been answered, which is critical knowledge for the system to track which goals remain open. In this example, the goal is explicitly closed in turn 3, indicated by the corresponding `CG' (close-goal) goal tracking annotation. Not all goals are explicitly closed by the user. A dialogue manager could decide to implicitly close an answered goal if the user does not reopen it within three turns, for example. We only annotate explicit goal closures, like the one in turn 3. There are two intents annotated in both turns 1 and 3 in Table \ref{tab:ex2}, and multiple goal annotations can similarly exist, separated by a semicolon. For example, ``I'm hungry but need the toilet first'' simultaneously opens two goals. All of these annotations were completed using the ELAN tool \cite{brugman2004annotating}, and then mapped into JSON for model training\footnote{Mapping code, annotated data, and training hyperparameters can be found here: \url{https://github.com/AddleseeHQ/mpgt-eval}.}.

With these two sets of annotations, we can evaluate various LLMs on two tasks: (1) multi-party intent-slot recognition; and (2) multi-party goal tracking. It is not possible to collect vast quantities of interactions with patients in the hospital, so these models must be able to learn from a corpus of limited size. We therefore decided to mask annotations in a randomised window selected from each MPC, providing the model with the surrounding context and speaker labels. That is, a random number of turns was selected in each MPC, and then the annotations were replaced by a `[MASK]' token. An example of this is shown in Table \ref{tab:ex3}.

As the corpus size is limited, the window selection could potentially heavily impact model performance. We therefore randomised the selected window three times for each conversation and train/test split, and these \emph{exact same} windows were used to train and test each model. To clarify, all train/test splits and windows were randomised for multiple runs, but they were unchanged between each model. For example, run 1 with the 20/80 split in Section \ref{sec:results} for T5 contained the exact same test set, with the exact same window, as run 1 with the 20/80 split for DialogLED. This holds true for both tasks. Each masked window was bookended with a\nobreakspace `[start]' and `[end]' tag to help the models learn this task too \cite{zhong2022dialoglm}. A shortened example from our corpus can be seen in Table \ref{tab:ex3}.

\begin{table*}
\begin{tabular}{llll}
\hline
  & \textbf{User} & \textbf{Masked Goal Tracking Utterance} & \textbf{Gold Annotation} \\ \hline 
1 & ARI: & Hello, my name is ARI. How can I help you? & - \\
  &      & [start] & - \\
2 & U1:  & My friend is intimidated by you, where can they eat? [MASK] & G(U2, eat())       \\
3 & ARI: & There's a cafeteria on the ground floor, near the courtyard. [MASK] & AG(U2, eat())       \\
  &      & [end] & - \\
4 & U2:  & My appointment is in room 17, where is it? G(U2, go-to(room\_17)) & -                              \\ \hline
\end{tabular}
\caption{A corpus example illustrating the goal tracking task. This process was the same for intent-slot recognition, with the corresponding annotations. Note that U1 asks U2's question, and this is reflected in the annotation.}
\label{tab:ex3}
\end{table*}

\section{Experimental Procedure}
\label{sec:experiment}

We evaluated three different models (each detailed below): T5 \cite{raffel2020exploring}, DialogLM using LED (DialogLED) \cite{zhong2022dialoglm}, and GPT-3.5-turbo\footnote{\url{https://platform.openai.com/docs/models/gpt-3-5}}. Each approach was evaluated in a zero-shot and few-shot setting, with various train/test splits. We could not provide more data to GPT-3.5-turbo due to context window size, but the train/test splits for T5 and DialogLED were: 0/100 (zero-shot), 20/80, 50/50, and 80/20. This allowed us to determine how each model learned to do these tasks when given more training examples. As described in Section \ref{sec:dataset}, we ran each experiment three times with randomised splits and windows, but these remained the same between-models to avoid few-shot problems such as  recency bias \cite{zhao2021calibrate}. We trained all the T5-Large and DialogLED models on a machine containing a 16Gb NVIDIA GeForce RTX 3080 Ti GPU with 64Gb RAM and an Intel i9-12900HK processor.

\subsection{T5-Large}
\label{sub:t5}

Older GPT models (GPT-3 and below) are pre-trained with the next token prediction objective on huge corpora \cite{radford2019language,brown2020language}, an inherently directional task. The creators of T5 added two more objectives and give it the goal of minimising the combined loss function \cite{raffel2020exploring} across all three tasks. The two additional tasks were de-shuffling, and BERT-style de-masking \cite{devlin2018bert}. This latter pre-training task involves `corrupting' tokens in the original text, which T5 must then predict. Importantly, this enabled T5 to work bidirectionally, becoming particularly good at using the surrounding context to predict tokens in corrupted sentences. This is not dissimilar to our task, in which the model must learn to use the surrounding MPC turns to predict the annotations that are masked. T5 also achieves state-of-the-art results on related tasks like \cite{lee2021dialogue,marselino2022comparative}, albeit, fine-tuned on larger datasets.

We used T5-Large in both a zero-shot setting, and fine-tuned with various train/test splits. T5 allows  fine-tuning with a given named task like `answer the question', or `translate from French to German'. We used `predict goals' and `predict intent-slots' for goal tracking and intent-slot recognition, respectively, giving the same task names as input during testing. As the corpus is very small, there was no model performance boost beyond 3 epochs, which was expected \cite{mueller2022label}.

\subsection{DialogLM using LED (DialogLED)}
\label{sub:dialogled}

MPCs reveal unique new communication challenges \cite{addlesee2023data}, as detailed in Section \ref{sec:intro}, so some LLMs have been developed specifically for the multi-party domain \cite{hu2019gsn,gu2021mpc,gu2022hetermpc}. Microsoft published DialogLM \cite{zhong2022dialoglm}, a pre-trained LLM based upon UniLMv2 \cite{bao2020unilmv2}, but specifically designed for multi-party tasks. Alongside the base model, they released two variations: DialogLM-sparse for long dialogues over 5,120 words, and DialogLM using LED (DialogLED) which outperformed the others. DialogLED builds on Longform-Encoder-Decoder (LED) \cite{beltagy2020longformer}, an attention mechanism that scales linearly with sequence length. Transformer-based models typically scale quadratically with the sequence length, restricting their ability to process long dialogues.

DialogLED was pre-trained on five objectives designed specifically for MPCs, and the model's goal was to minimise the combined loss of all of these tasks. Their state-of-the-art results showed that their pre-training tasks did encourage the LLM to `understand' multi-party interactions. The five tasks were: (1) speaker masking, the model has to predict who spoke; (2) turn splitting, the model has to recognise when two utterances are likely the same turn; (3) turn merging, the opposite of (2), where the model has to recognise when the turns were likely separate; (4) text infilling, the model has to predict masked tokens within the turn; and (5) turn permutation, the model has to correctly re-order  jumbled turns.

We cloned their repository\footnote{\url{https://github.com/microsoft/DialogLM}} and added two new tasks: (6) goal masking, the model has to predict goal tracking annotations; and (7) intent-slot masking, the model has to predict intent-slot annotations. In the zero-shot setting, we simply ran the test set through base DialogLED. We then ran their, now modified, code to run our few-shot evaluations three times for each data split.

\subsection{GPT-3.5-turbo}
\label{sub:gpt}

Larger LLMs are not inherently better at following a user's intent \cite{ouyang2022training} as they have no incentive to help the user achieve their goal, only to generate realistic looking outputs. This leads to significant problems, including the generation of false, biased, and potentially harmful responses. GPT-3 was therefore fine-tuned on prompts with human-feedback to create InstructGPT \cite{ouyang2022training}. OpenAI later followed this same approach to create the now famous ChatGPT family of models. At the time of writing, GPT-4 is the most powerful of these models, but it is currently in a waiting list phase. OpenAI recommends their GPT-3.5-turbo model while waiting as the next best option. We therefore decided to evaluate this model on the same two tasks.

Unlike T5 or DialogLED, there is no way to fine-tune your own version of GPT-3.5-turbo, or to edit their pre-training steps. People instead mould the model's behaviour through prompt-engineering \cite{lester2021power,wei2022chain,weng2023prompt}. The newer GPT models allow developers to provide huge contexts, called prompts, containing instructions for the model to follow. GPT-3.5-turbo allows prompts of up to 4,096 tokens. Although these models have only exploded in popularity recently, there are many suggested prompt `styles' suggested online by conversation designers who are implementing these models in the real-world. We have analysed this space and devised six prompt styles for the two tasks. In the zero-shot setting, only the prompt and the masked MPC is provided to the model. In the few-shot setting, we additionally provide the model with 7\% of the corpus as examples. This is crucial to highlight. T5 and DialogLED were trained on 20\% of the corpus, 50\% of the corpus, and finally 80\% of the corpus. GPT-3.5-turbo's maximum context size can only fit 7\% of the corpus, less than the other models.

The prompt styles we used were the following (the actual prompts are included in Appendix \ref{sec:prompts}):
\begin{itemize}
    \item \textbf{Basic}: This is our baseline prompt. It very simply tells the model what it is going to get as input, and what we want as output. It contains no further special instructions.
    \item \textbf{Specific}: GPT practitioners report that when   prompts are more detailed and specific, performance is  boosted \cite{ye2023context}.
    \item \textbf{Annotation}: For annotation tasks, we would give fellow humans annotation instructions. In this prompt, we provide the model with annotation instructions.
    \item \textbf{Story}: This model was pre-trained on a very large quantity of data, including novels, film scripts, journalistic content, etc... It may be possible that by phrasing the prompt like a\nobreakspace story, performance may be boosted due to its likeness to its training data.
    \item \textbf{Role-play}: Similar to the story prompt, it is reported that these models are very good at role-playing\footnote{\url{https://github.com/f/awesome-chatgpt-prompts}}. People ask ChatGPT to pretend to be a therapist, a lawyer, or even alter-egos that have no safety limitations \cite{taylor2023chatGPT}. We tell GPT-3.5-turbo that it is a `helpful assistant listening to a conversation between two people and a social robot called ARI'.
    \item \textbf{Reasoning}: Finally, recent work suggests that these models improve in performance if you explain the reasoning for desired outputs \cite{fu2022complexity}. We therefore added one fictitious turn to this prompt, and explained the reasoning behind its annotation.
\end{itemize}

\section{Results}
\label{sec:results}

We evaluated T5, DialogLED, and GPT-3.5-turbo as described in Section \ref{sec:experiment} on multi-party goal tracking, and multi-party intent-slot recognition. Outputs were annotated as either `exact', `correct', or `partial' to distinguish each model's performance beyond simple accuracy. Exact matches were strictly annotated, but slight differences are allowed if the annotation meaning remains unchanged. For example: `{\tt G(U1, go-to(lift))}' and `{\tt G(U1, go-to(lifts))}' (note the plural `lifts'). Outputs were marked as exact if every [MASK] in the MPC was exact, and marked as correct if every [MASK] was more broadly accurate. For example, if the annotation contained `{\tt drink(coffee)}' and the model output `{\tt drink(hot\_drink)}', we considered this correct. The output was marked as partially correct if at least 60\% of the [MASK] tags were correctly annotated. This latter metric allows us to distinguish between models that generate nonsense, and those that roughly grasp the task. Our inter-annotator agreements were 0.765 and 0.771 for goal tracking and intent-slot recognition, respectively. These are less than 0.8, and this was due to the broad definition of `correct'. We plan to design automatic metrics for our future work (see Section \ref{sec:conc}).

\subsection{MPC Goal Tracking Results}

The goal tracking results can be found in Table \ref{tab:goal-only-results}. An ANOVA test \cite{fisher1992statistical} indicated that there was an overall significant difference between the model's results. We therefore ran a Tukey HSD test \cite{tukey1949comparing} that showed that the GPT-3.5-turbo model in the few-shot setting did significantly outperform all the other models.

\begin{table*}
\centering
\begin{tabular}{|ccc|lll|}
\hline
\textbf{Model}  & \textbf{train/test \%} & \textbf{Prompt Style} & \textbf{Exact \%}    & \textbf{Correct \%}         & \textbf{Partial \%}                             \\
\hline
T5            & 0/100         & -            & 0                                      & 0                                      & 0                                      \\
T5            & 20/80         & -            & 0 {\footnotesize $\pm$ 0}                 & 0 {\footnotesize $\pm$ 0}                 & 0 {\footnotesize $\pm$ 0}                 \\
T5            & 50/50         & -            & 0 {\footnotesize $\pm$ 0}                 & 0 {\footnotesize $\pm$ 0}                 & 0 {\footnotesize $\pm$ 0}                 \\
T5            & 80/20         & -            & 0 {\footnotesize $\pm$ 0}                 & 0 {\footnotesize $\pm$ 0}                 & 0 {\footnotesize $\pm$ 0}                 \\
\hline \hline
DialogLED     & 0/100         & -            & 0                                      & 0                                      & 0                                      \\
DialogLED     & 20/80         & -            & 0 {\footnotesize $\pm$ 0}                 & 0 {\footnotesize $\pm$ 0}                 & 5.80 {\footnotesize $\pm$ 1.45}           \\
DialogLED     & 50/50         & -            & 0 {\footnotesize $\pm$ 0}                 & 2.38 {\footnotesize $\pm$ 2.38}           & 1.19 {\footnotesize $\pm$ 0.63}           \\
DialogLED     & 80/20         & -            & 0 {\footnotesize $\pm$ 0}                 & 0 {\footnotesize $\pm$ 0}                 & 20 {\footnotesize $\pm$ 11.55}            \\
\hline \hline
GPT 3.5-turbo & 0/100         & Basic        & 0                                      & 3.45                                   & 31.03                                  \\
GPT 3.5-turbo & 0/100         & Specific     & 0                                      & 3.45                                   & 24.14                                  \\
GPT 3.5-turbo & 0/100         & Annotation   & 0                                      & 6.90                                   & 44.83                                  \\
GPT 3.5-turbo & 0/100         & Story        & 0                                      & 0                                      & 0                                      \\
GPT 3.5-turbo & 0/100         & Role-play     & 0                                      & 0                                      & 6.90                                   \\
GPT 3.5-turbo & 0/100         & Reasoning    & 3.45                                   & 34.48                                  & 79.31                                  \\
\hline
GPT 3.5-turbo & 7/80*        & Basic        & 11.59 {\footnotesize $\pm$ 3.83}          & 30.43 {\footnotesize $\pm$ 10.94}         & 86.96 {\footnotesize $\pm$ 6.64}          \\
GPT 3.5-turbo & 7/80*        & Specific     & 20.29 {\footnotesize $\pm$ 3.83}          & 43.48 {\footnotesize $\pm$ 9.05}          & 92.75 {\footnotesize $\pm$ 2.90}          \\
GPT 3.5-turbo & 7/80*        & Annotation   & 14.49 {\footnotesize $\pm$ 5.80}          & 28.99 {\footnotesize $\pm$ 3.83}          & 82.61 {\footnotesize $\pm$ 4.35}          \\
GPT 3.5-turbo & 7/80*        & Story        & 17.39 {\footnotesize $\pm$ 6.64}          & 36.23 {\footnotesize $\pm$ 13.83}         & 86.96 {\footnotesize $\pm$ 4.35}          \\
GPT 3.5-turbo & 7/80*        & Role-play     & 18.84 {\footnotesize $\pm$ 7.25}          & 46.38 {\footnotesize $\pm$ 12.38}         & 92.75 {\footnotesize $\pm$ 5.22}          \\
GPT 3.5-turbo & 7/80*        & Reasoning    & \textbf{27.54 {\footnotesize $\pm$ 1.45}} & \textbf{62.32 {\footnotesize $\pm$ 9.50}} & \textbf{94.20 {\footnotesize $\pm$ 5.80}} \\
\hline
\end{tabular}
\caption{The final multi-party goal tracking results for each model in both the zero- and few-shot settings. *We\nobreakspace could\nobreakspace not\nobreakspace fit more than 7\% of the training examples in GPT-3.5-turbo's context window. We therefore used fewer examples than with T5 and DialogLED. The same 80\% test sets were still used to enable model comparison.}
\label{tab:goal-only-results}
\end{table*}

Firstly, the T5-Large model performed poorly, even when it was trained on 80\% of our corpus. Upon further analysis, it generated complete nonsense in the zero-shot setting, but did start to generate strings that looked reasonable with only 20\% of the data. Given the 50/50 train/test split, T5 consistently replaced the [MASK] tokens, but did still hallucinate turns. When given 80\% of the data as training data, the T5 model preserved the original dialogue, and replaced the [MASK] tokens with goal annotations, they were just all completely wrong. This steady improvement as we increased the amount of training data suggests that T5 could be a viable option for similar tasks, just not where data is limited (such as our hospital use case).

The DialogLED model also generated nonsense in the zero-shot setting, but very quickly learned the task. Even with just 20\% of the data used for training, DialogLED reliably preserved the original dialogue and replaced the [MASK] tokens with goal annotations. Most of the annotations were incorrect, for example `{\tt G(U2, eat(ticket))}', but DialogLED did correctly detect some goals opening, being answered, and being closed correctly, achieving a non-zero partial score. Given more training data, DialogLED did begin to use the surrounding contextual dialogue turns more accurately, but almost every result contained an incorrect prediction. This was often the mis-detection of shared goals, or closing goals early. Like T5, DialogLED would need a larger training set to accurately complete this task. This model learned the task quickly, so may need fewer examples.

In the zero-shot setting, GPT-3.5-turbo roughly `understood' the task, generating many partially correct outputs. With all the prompt styles, it did frequently reformat the dialogue. This was particularly true when using the roleplay prompt, it would output all the goals per interlocutor, for example, rather than per turn. The worst zero-shot GPT-3.5-turbo prompt was the `story' style, not even generating one partially correct output. This was due to its increased hallucination. The story prompt noticeably produced more fictitious turns, and also rephrased and removed turns in the original dialogue. We believe this is likely because a story scenario is naturally a fictitious topic. The `reasoning' style prompt performed remarkably well, generating five times more correct outputs than the second-best prompt style, and generating 79.31\% partially correct outputs, showing that it can grasp the concept of the task. The reasoning prompt commonly mis-identified shared goals, unfortunately.

In the few-shot setting, GPT-3.5-turbo's results improved significantly compared to every other approach. We would like to highlight again that each run's example prompts provided to the model were exactly the same for each prompt style. Performance differences were only due to the given prompt style. The `reasoning' prompt once again outperformed the others across all metrics, generating correct outputs 62.32\% of the time, and partially correct 94.20\% of the time. In our future work (see Section \ref{sec:conc}), we plan to utilise this prompt style's impressive performance on limited data. The `story' prompt was the only style to successfully attribute goals to other speakers, as in Table \ref{tab:ex3}, but it still suffered from increased hallucination, which is not appropriate in a safety-critical setting. We suspect that the other prompt styles failed to do this because of the rarity of this phenomenon in our corpus. We are eliciting more of these in ongoing experiments with a deployed system, not wizard-of-oz \cite{addlesee2023data}.

\subsection{MPC Intent-slot Recognition Results}

The results for each model on the intent-slot recognition task can be found in Table \ref{tab:dst-only-results}. As with the goal tracking results, an ANOVA test \cite{fisher1992statistical} indicated that there was an overall significant difference between our model's results. We therefore ran a Tukey HSD test \cite{tukey1949comparing} that showed that the GPT-3.5-turbo model in the few-shot setting significantly outperformed all the other models.

\begin{table*}
\centering
\begin{tabular}{|ccc|lll|}
\hline
\textbf{Model}  & \textbf{train/test \%} & \textbf{Prompt Style} & \textbf{Exact \%}    & \textbf{Correct \%}         & \textbf{Partial \%}                             \\
\hline
T5            & 0/100         & -            & 0                                      & 0                                      & 0                                      \\
T5            & 20/80         & -            & 0 {\footnotesize $\pm$ 0}                 & 0 {\footnotesize $\pm$ 0}                 & 0 {\footnotesize $\pm$ 0}                 \\
T5            & 50/50         & -            & 0 {\footnotesize $\pm$ 0}                 & 0 {\footnotesize $\pm$ 0}                 & 0 {\footnotesize $\pm$ 0}                 \\
T5            & 80/20         & -            & 0 {\footnotesize $\pm$ 0}                 & 0 {\footnotesize $\pm$ 0}                 & 0 {\footnotesize $\pm$ 0}                 \\
\hline \hline
DialogLED     & 0/100         & -            & 0                                      & 0                                      & 0                                      \\
DialogLED     & 20/80         & -            & 0 {\footnotesize $\pm$ 0}                 & 0 {\footnotesize $\pm$ 0}                 & 5.80 {\footnotesize $\pm$ 2.90}           \\
DialogLED     & 50/50         & -            & 0 {\footnotesize $\pm$ 0}                 & 0 {\footnotesize $\pm$ 0}           & 38.10 {\footnotesize $\pm$ 10.38}           \\
DialogLED     & 80/20         & -            & 0 {\footnotesize $\pm$ 0}                 & 13.33 {\footnotesize $\pm$ 6.67}                 & 46.67 {\footnotesize $\pm$ 6.67}            \\
\hline \hline
GPT 3.5-turbo & 0/100         & Basic        & 0                                      & 3.45                                   & 51.72                                  \\
GPT 3.5-turbo & 0/100         & Specific     & 0                                      & 0                                   & 13.79                                  \\
GPT 3.5-turbo & 0/100         & Annotation   & 0                                      & 3.45                                   & 20.69                                  \\
GPT 3.5-turbo & 0/100         & Story        & 0                                      & 0                                      & 24.14                                      \\
GPT 3.5-turbo & 0/100         & Role-play     & 0                                      & 0                                      & 20.69                                   \\
GPT 3.5-turbo & 0/100         & Reasoning    & 0                                   & 27.59                                  & 82.76                                  \\
\hline
GPT 3.5-turbo & 7/80*        & Basic        & 17.39 {\footnotesize $\pm$ 6.64}          & 36.23 {\footnotesize $\pm$ 12.88}         & 97.10 {\footnotesize $\pm$ 2.90}          \\
GPT 3.5-turbo & 7/80*        & Specific     & 27.54 {\footnotesize $\pm$ 1.45}          & 60.87 {\footnotesize $\pm$ 9.05}          & 94.20 {\footnotesize $\pm$ 1.45}          \\
GPT 3.5-turbo & 7/80*        & Annotation   & 18.84 {\footnotesize $\pm$ 1.45}          & 40.58 {\footnotesize $\pm$ 6.32}          & 91.30 {\footnotesize $\pm$ 4.35}          \\
GPT 3.5-turbo & 7/80*        & Story        & 26.09 {\footnotesize $\pm$ 4.35}          & 47.83 {\footnotesize $\pm$ 10.04}         & 94.20 {\footnotesize $\pm$ 3.83}          \\
GPT 3.5-turbo & 7/80*        & Role-play     & 20.29 {\footnotesize $\pm$ 3.83}         & 49.27 {\footnotesize $\pm$ 12.88}         & 97.10 {\footnotesize $\pm$ 1.45}          \\
GPT 3.5-turbo & 7/80*        & Reasoning    & \textbf{37.68 {\footnotesize $\pm$ 1.45}} & \textbf{69.57 {\footnotesize $\pm$ 10.94}} & \textbf{100 {\footnotesize $\pm$ 0}} \\
\hline
\end{tabular}
\caption{The final multi-party intent-slot recognition results for each model in both the zero- and few-shot settings. *We could not fit more than 7\% of the training examples in GPT-3.5-turbo's context window. We therefore used fewer examples than with T5 and DialogLED. The same 80\% test sets were still used to enable model comparison.}
\label{tab:dst-only-results}
\end{table*}

As intent-slot annotations are well-established, T5 and DialogLED both started generating sensible-\emph{looking} outputs with only a few training examples. The T5 outputs were all incorrect again, however. DialogLED consistently improved as it was trained on progressively more data, annotating almost half of the MPCs partially correctly, and beginning to accurately annotate full MPCs. Given a larger corpus, we expect that DialogLED could potentially generate competitive results, but this is not the case for T5 in this setting with limited data.

GPT-3.5-turbo in the zero-shot setting also achieved higher partial scores, compared to the goal tracking results, due to the fact that intent-slot recognition is a more established task. Turns were commonly annotated with multiple gold goals, but this model tended to only output one per turn. For example: ``Hello ARI, where is the café?'' would only have the prediction `greet', missing the request to locate the café entirely. This prevented the model from achieving higher correct scores.

In the few-shot setting, however, GPT-3.5-turbo significantly outperformed all the other models. The difference was remarkable. Almost all of the predictions were partially correct, and the `reasoning' prompts correctly annotated 70\% of the MPCs. Other models tended to falter when anaphoric expressions couldn't be resolved with just the previous turn. They also struggled to identify the `{\tt suggest}' intent, for example, when one person said ``do you want to go to the toilet?''. These were misclassified as request intents, likely due to their prominence in the corpus, and influence on the results due to GPT-3.5-turbo's limited input context.



\section{Conclusion and Future Work}
\label{sec:conc}

Multi-party conversations (MPCs) elicit complex behaviours which do not occur in the dyadic interactions that today's dialogue systems are designed and trained to handle. Social robots are increasingly being expected to perform tasks in public spaces like museums and malls, where conversations often include groups of friends or family. Multi-party research has previously focused on speaker recognition, addressee recognition, and tweaking response generation depending on whom the system is addressing. While this work is vital, we argue that these collective ``who says what to whom'' tasks do not provide any incentive for the social robot to complete user goals, and instead encourage it to simply mimic what a good MPC \emph{looks like}. In this paper, we have detailed how the tasks of goal tracking and intent-slot recognition differ in a multi-party setting, providing examples from our newly collected corpus of MPCs in a hospital. We found that, given limited data, `reasoning' style prompts enable GPT-3.5-turbo to perform significantly better than other models.

We found that other prompt styles also perform well, but prompts that are story-like increase model hallucination. With the introduction of prompt fine-tuning with human feedback \cite{ouyang2022training}, generative LLMs do now have some incentive to avoid misleading or harming the user, providing outputs prepended with caveats, but the issue is not solved. OpenAI claims that GPT-4 generates 40\% fewer hallucinations than GPT-3 \cite{hern2023openAI}, but these models should still not be applied directly in a hospital or other safety-critical  setting without further evaluation. In the hospital setting, users are more likely to be from vulnerable population groups, and are more likely to be older adults that are not familiar with the capabilities of today's models. Multiple researchers and hospital staff members are present when conducting our data collections, so that if hallucinations do occur, they can be quickly corrected. We will, therefore, be able to evaluate response grounding, Guidance\footnote{\url{https://github.com/microsoft/guidance}}, and other hallucination prevention strategies to determine whether these models can ever be used safely in a high-risk setting. These further experiments will also elicit further MPCs that can be annotated for various multi-party tasks.

User inputs must be processed on external servers when using industry LLMs, like GPT-3.5-turbo and Google's Bard. For this reason, these specific models cannot be deployed in the hospital setting. Patients may reveal identifiable or sensitive information during our data collection, which we subsequently remove from the corpus. This data must stay contained within approved data-controlled servers in the SPRING project. In this paper, we have reported the remarkable performance of an industry LLM, when given limited data, compared to prior model architectures. We will analyse open and transparent instruction-tuned text generators \cite{liesenfeld2023opening}, which are able to meet our data security requirements.

The accessibility of today's SDSs is critical when working with hospital patients \cite{addlesee2023voice}. Speech production differs between the `average' user, and user groups that remain a minority in huge training datasets. For example, people with dementia pause more frequently and for longer durations mid-sentence due to word-finding problems \cite{boschi2017connected, slegers2018connected}. We are utilising knowledge graphs to ensure that SDSs are transparent, controllable, and more accessible for these user groups \cite{addlesee-eshghi-2021-incremental,addlesee2023understandingsparql,addlesee2023understandingamr}, and we see the unification of large language models and knowledge graphs \cite{pan2023unifying} as the near-term future of our field.

We plan to design and run subsequent experiments in both the hospital memory clinic, and a newly established mock waiting room in our lab. This space will allow us to collect additional MPCs with more than two people, replicating scenarios in which whole families approach a social robot. We plan to evaluate whether prompt engineering can work modularly for N users. For example, we could use GPT-4 to correct speaker diarization \cite{murali2023improving}, then to handle multi-party goal tracking, and then to generate responses to the user. This experimental setup will allow us to quickly test new ideas, such as automatic prompt optimization \cite{pryzant2023automatic} in the lab, maximising the benefit of patients' time in the hospital.

\section*{Acknowledgements}

This research was funded by the EU H2020 program under grant agreement no. 871245 (\url{https://spring-h2020.eu/}). We would also like to thank our anonymous reviewers for their time and valuable feedback.

\bibliography{anthology,custom}
\bibliographystyle{acl_natbib}

\appendix

\section{Full GPT-3.5-turbo Prompts}
\label{sec:prompts}

Here are the full prompts given to GPT-3.5-turbo for each task. We used six styles described in Section \ref{sec:experiment}. The masked MPC was appended to each prompt in the zero-shot setting. In the few-shot prompts (see Section \ref{sub:few}), we appended examples with ``input:'' + masked MPC \#1 + ``output:'' + gold output \#1 + `input:'' + masked MPC \#2 + ``output:'' + gold output \#2 + ``input:'' + test set masked MPC + ``output:''\footnote{The examples given were randomised per run, and the appendix page limit doesn't fit the full 4,096 token prompts.}.

\subsection{Zero-shot Goal Tracking}

\begin{itemize}
    \item \textbf{Basic}: This conversation has a window between [start] and [end]. Return this window with the [MASK] tags replaced with the goal annotations:
    \item \textbf{Specific}: This is a conversation between two people and a robot called ARI. There is a section of the conversation between the [start] and [end] tags. I want you to return this section of the conversation, but I want you to replace the [MASK] tags with the user goals. Do not change any of the other words in the section, only replace [MASK]. Every [MASK] should be replaced. Here is the conversation:
    \item \textbf{Annotation}: This is a conversation between two people and a robot called ARI. I want you to first extract the text between [start] and [end]. There are [MASK] tags in the extracted text. I want you to replace the [MASK] tags with goal annotations. Do not change any of the other text. If the person's goal can be determined by that turn, add an '@' symbol followed by 'G' (G for goal), and then brackets with the speaker ID and what their goal is. If it is a shared goal, you can annotate both speakers with a '+' sign between them. For example, if you think U1 and U2 share the goal, you can write U1+U2. If you think the goal is being answered, you can do the same but with 'AG' (AG for Answer Goal) instead of 'G'. Finally, if you think the person is closing the goal, you can do the same annotation using 'CG' (CG for Close Goal) instead of 'G' or AG'. Here is the conversation:
    \item \textbf{Story}: There once was a conversation between a patient, a companion, and a robot called ARI. One bit of the conversation was confusing. A helpful researcher noted the start with [start], and the end with [end]. The confusing bits are marked with [MASK]. Can you help us figure out the goals that should replace the [MASK] tags? The conversation is this:
    \item \textbf{Role-play}: You are listening to a conversation between two people and a robot called ARI. You are a helpful assistant that needs to figure out what goals the people have. You need to pay attention to the [MASK] tags between the [start] and [end] tags in the given conversation. Your job is to replace these [MASK] tags with the correct goal annotations. Here is the conversation:
    \item \textbf{Reasoning}: I will give you a conversation between two people and a robot called ARI. You need to return the text between [start] and [end] with the [MASK] tags replaced by user goals. Let's step through how to figure out the correct annotation. If the conversation included 'U1: I really need the toilet [MASK]', then we would first know that the speaker is called U1. The turn also ends with [MASK], so we know that we need to replace it with a goal. We know that U1 needs the toilet, so their goal is to go to the nearest toilet. Goals always begin with the '@' symbol, and then a 'G' if we have found a person's goal. We would therefore replace [MASK] with @ G(U1, go-to(toilet)). If someone tells U1 where the toilets are, they have answered their goal. We would therefore annotate that turn with @ AG(U1, go-to(toilet)). We use AG here to indicate Answer Goal. Finally, if U1 then said thank you, we know their goal has been met. We would annotate the thank you with @ CG(U1, go-to(toilet)) because U1's goal is finished. CG stands for Close Goal. Do this goal tracking for each [MASK] in this conversation:
\end{itemize}

\subsection{Few-shot Intent-slot Recognition}
\label{sub:few}

\begin{itemize}
    \item \textbf{Basic}: Each conversation has a window between [start] and [end]. Return this window with the [MASK] tags replaced with the intent-slot annotations. Here are some examples.
    \item \textbf{Specific}: Each of these conversations is between two people and a robot called ARI. There is a section of each conversation between the [start] and [end] tags. I want you to return this section of the conversation, but I want you to replace the [MASK] tags with the user intents and slots. Do not change any of the other words in the section, only replace [MASK]. Every [MASK] should be replaced. Here are some examples.
    \item \textbf{Annotation}: Each of these conversations is between two people and a robot called ARI. I want you to first extract the text between [start] and [end]. There are [MASK] tags in the extracted text. I want you to replace the [MASK] tags with intent-slot annotations. Do not change any of the other text. If the person's intent can be determined by that turn, add a '\#' symbol followed by their intent and then brackets with the slots within. There are not always slots, so the brackets can be empty. Sometimes there are multiple intents, split them with a semi-colon ';'. Here are some examples.
    \item \textbf{Story}: There once was a conversation between a patient, a companion, and a robot called ARI. One bit of the conversation was confusing. A helpful researcher noted the start with [start], and the end with [end]. The confusing bits are marked with [MASK]. Can you help us figure out the intents and slots that should replace the [MASK] tags? Here are some examples.
    \item \textbf{Role-play}: You are listening to a conversation between two people and a robot called ARI. You are a helpful assistant that needs to figure out what goals the people have. You need to pay attention to the [MASK] tags between the [start] and [end] tags in the given conversation. Your job is to replace these [MASK] tags with the correct intent-slot annotations. Here are some examples.
    \item \textbf{Reasoning}: I will give you a conversation between two people and a robot called ARI. You need to return the text between [start] and [end] with the [MASK] tags replaced by user intents and slots. Let's step through how to figure out the correct annotation. If the conversation included 'U1: Hello, I'd like to know where the doctor's office is? [MASK]' then we know there is a missing intent-slot annotation because of the [MASK] tag. U1 first said hello, greeting their interlocutor, so we know their intent is greet. This has no slots, so we have the annotation '\# greet()' to start. U1 also asked where the doctor is, so their second intent is a request. The slot is the room that the doctor is in, as that is what they are requesting. Their second intent is therefore '\# request(doctor(room)). As there are multiple intents, the [MASK] is replaced by '\# greet() ; request(doctor(room))'. The ';' is only used because there was more than one intent. Do this intent-slot annotation for each [MASK] in this conversation. Here are some examples.
\end{itemize}

\end{document}